\documentclass[11pt]{article}
\usepackage[margin=1.1in]{geometry}
\usepackage[T1]{fontenc}
\usepackage[utf8]{inputenc}
\usepackage{lmodern}
\usepackage{microtype}
\usepackage{booktabs}
\usepackage{array}
\usepackage[square,semicolon,authoryear]{natbib}
\usepackage[hidelinks]{hyperref}
\usepackage{orcidlink}
\usepackage{url}
\hypersetup{pdftitle={MedWER: A Reproducible, Model-Free Evaluation Protocol
    for Medical Speech Recognition},
  pdfauthor={Justin Behling},
  pdfsubject={An evaluation protocol and open-source tool for medical ASR
    with a fixed, model-free term-list denominator},
  pdfkeywords={speech recognition, medical ASR, word error rate,
    evaluation metric, reproducibility}}

\title{MedWER: A Reproducible, Model-Free Evaluation Protocol\\for Medical
  Speech Recognition\thanks{Code, term list, and golden fixtures:
    \url{https://github.com/Nordis-Tech/medwer}.}}
\author{Justin Behling\,\orcidlink{0000-0002-6491-0391}\\
  NORDIS TECH INC.\\
  \texttt{justin.behling@nordistech.ca}}
\date{}

\newcommand{\medwer}{medical-WER}

\begin{document}
\maketitle

\begin{abstract}
Overall word error rate hides clinically critical errors: a transcript can be 95\%
correct and still swap one drug for another. The usual fix weights errors on
medical entities, and almost always depends on an evaluation-time
named-entity recognition (NER) model or cloud API, which makes the metric's
denominator a versioned black box. We present
\textbf{MedWER}, an evaluation protocol and open-source tool for medical ASR
whose denominator is a \emph{fixed, license-clean term list}: 19{,}373 drug,
diagnosis, symptom, and injury-mechanism entries projected from public
sources. The protocol couples a pinned text normalizer with a phrase-aware
term-restricted WER, the \medwer{}, so the only versioned component is a
normalizer dependency held at an exact release and checked against committed
golden fixtures. Coverage is validated against an independent provincial
drug-benefit file the list was not built from; the matching heuristic is
calibrated against ground-truth entity spans. Baselines for Moonshine~base,
Whisper~base.en, and MedASR on two open benchmarks are scored with the
released tool and reported with 95\% confidence intervals from resampled
per-utterance scores.
\end{abstract}

\section{Introduction}
\label{sec:intro}

Overall word error rate (WER) under-weights exactly the words that matter in
clinical dictation. PriMock57, a conversational medical benchmark, is 2.70\%
medical terms by token count (Section~\ref{sec:baselines}), so a system can
post an attractive WER while failing precisely on the drug and diagnosis
vocabulary.
This observation is not new, and a
family of entity-weighted metrics has grown up around it
(Section~\ref{sec:related}). What the family shares is a dependency: the set
of ``medical'' tokens is discovered at evaluation time by a named-entity
recognition (NER) model or a commercial API. That choice couples the metric's
denominator to a moving, often closed artifact. Two labs running the same
hypothesis files can get different medical error rates; the same lab can get
different numbers a year apart.

MedWER inverts the dependency. The denominator is a \emph{fixed artifact}: a
term list, versioned and redistributed with the tool, projected from public
sources that permit open commercial redistribution.
The metric machinery on top of it is plain: a pinned
normalizer and token-level Levenshtein WER restricted to term-list content.
The contributions are operational rather than conceptual:

\begin{enumerate}
\item \textbf{A model-free, deterministic denominator.} Term membership is
  string matching against a fixed list; multi-word terms
  are matched greedily longest-first and scored as single atomic units
  (Section~\ref{sec:protocol}).
\item \textbf{Coverage validated against an independent holdout.} The list's
  completeness is measured against a provincial drug-benefit file deliberately
  \emph{not} used in its construction (Section~\ref{sec:coverage}); the matching heuristic is
  calibrated against human entity-span annotations
  (Section~\ref{sec:calibration}).
\item \textbf{A pinned, fixture-checked normalizer.} The normalization
  dependency is held at an exact release, and its behavior is asserted
  against committed golden fixtures by the test suite, so an upstream change
  fails the build rather than silently moving published numbers
  (Section~\ref{sec:parity}).
\item \textbf{License-clean provenance end to end.} Every source in the term
  list permits redistribution in a commercial open-source artifact; the
  provenance is documented per source and ships with the data
  (Section~\ref{sec:vocab}).
\end{enumerate}

\section{Related work}
\label{sec:related}

\paragraph{Entity-weighted ASR metrics.} The closest prior work is
\citet{afonja2024performant}, who evaluate ASR on accented clinical speech
with a medical-WER, medical named-entity recall, and character error rate,
using fuzzy entity alignment, and demonstrate large fine-tuning gains on
accented clinical entities. A broader cluster shares the same motivation:
clinical-concept WER variants~\citep{adedeji2024sound}, entity error
rates~\citep{deng2026speechllm}, and semantic distance
measures~\citep{kim2021semdist}. In virtually all such metrics the medical or
entity content is identified at evaluation time by an NER tagger, annotation
model, or embedding model.

\paragraph{Reproducible ASR benchmarking.} The Open ASR
Leaderboard~\citep{srivastav2025openasr} established standardized text
normalization and joint reporting of WER and inverse real-time factor as the
baseline discipline for comparable ASR evaluation, with a shared
normalization pipeline following Whisper's \citep{radford2023whisper}.
MedWER adopts that normalizer unmodified and pinned, and extends the
reproducibility discipline in one direction: a fixed domain denominator that no
model computes.

\paragraph{Models and benchmarks used in this paper.} The systems evaluated
here are Moonshine~base~\citep{jeffries2024moonshine}, Whisper~base.en run under
\texttt{whisper.cpp} \citep{radford2023whisper}, and
MedASR~\citep{medasr2026}, a Conformer trained for physician dictation (105M
parameters and $\sim$5{,}000 hours per its HAI-DEF model card). The two public
sets are the Eka
medical ASR evaluation set~\citep{ekacare2025dataset} (real accented English
with real drug vocabulary) and PriMock57~\citep{korfiatis2022primock57}
(mock primary-care consultations; conversational register).

\section{The typed vocabulary and the term list}
\label{sec:vocab}

\subsection{Sources and licensing}

MedWER's term list is projected from a typed vocabulary built exclusively
from sources whose licenses permit open commercial redistribution of derived
term strings:

\begin{table}[ht]
\centering
\small
\begin{tabular}{l>{\raggedright\arraybackslash}p{1.38in}l}
\toprule
Term type & Source & License basis \\
\midrule
drug & Health Canada Drug Product Database (DPD) & Open Government Licence -- Canada \\
     & Canadian Clinical Drug Data Set (CCDD; Canada Health Infoway) & Open Government Licence -- Canada \\
diagnosis / symptom / mechanism & ICD-10-CM (NCHS/CMS) & U.S. Government work; public domain \\
\bottomrule
\end{tabular}
\caption{Term-list sources.}
\label{tab:sources}
\end{table}

Several otherwise-suitable sources were rejected on licensing grounds.
ICD-10-CA is CIHI-licensed and non-redistributable; the CM-versus-CA
distinction is code structure, not spoken nouns. ICD-11 is CC~BY-ND, and a
tokenized term list is a prohibited derivative. SNOMED~CT is licensed. LOINC,
the candidate for laboratory terms, is oversized, method-coded, and carries
attribution and license-notice obligations we chose not to propagate.
Laboratory and organism vocabularies are consequently absent: no open
structured source for either survived these constraints.

\subsection{Curation}

The critical observation for the drug type is that DPD/CCDD list everything
\emph{regulated} as a drug in Canada, including sunscreens, disinfectants,
veterinary products, and cosmetics: in scope for the source, but not dictated
clinical drugs. These are removed structurally rather than by name-level
judgment: the DPD \texttt{CLASS} field (keep \emph{Human}; drop Disinfectant,
Veterinary, Radiopharmaceutical), a non-therapeutic ATC cut, and a keyword
denylist for the residue. The cuts leave 13{,}672 drug records from 15{,}994
raw.
ICD-10-CM is ingested once and routed by chapter into diagnosis (A--Q, S--T),
symptom (R), and injury-mechanism (V--X, Y00--Y38) types, then cleaned from
coding strings toward dictation phrases. The cleaning steps are parenthetical
and bracket stripping, comma truncation, truncation at a small set of
clause-boundary tokens, and rejection of coding filler, laterality variants,
and over-long strings; they yield 6{,}048 / 332 / 327 records respectively.
Every surviving record therefore traces to a field of DPD/CCDD or to an
ICD-10-CM chapter; the filtering rules above are the whole of the editorial
input.

\subsection{Projection}

The published term list is a projection of this vocabulary: 19{,}373 entries,
11{,}742 of them multi-word, the longest 15 tokens. Every entry is projected
through the protocol normalizer to a fixed point of it, so entries match
normalized text verbatim.
Multi-word names are carried as whole phrase lines; their components of at
least four characters that fall outside a 10{,}000-word common-English list
are additionally emitted as single-token entries. Brand product names
contribute product-line words by the same rule (\emph{bedtime},
\emph{forte}). Single-word source names enter as themselves and bypass those two
filters, so the length and common-word rules bound only what decomposition
adds. One rule applies to single-token entries from either path: an entry must
contain a letter. A bare strength or product code (\emph{0.5\%}, \emph{12201})
identifies nothing on its own, and number folding would make it match every
spoken number in a reference; the multi-word names such tokens come from are
still carried whole.

Phrase-awareness changes what the denominator \emph{contains}. Under a
single-token list a multi-word drug or diagnosis name contributes only its
component words; phrase lines admit the whole terms, which raises \medwer{}
on jargon-bearing material with recognition unchanged.
The same machinery covers spelled-letter entries (\texttt{c h f}) for corpora
that transcribe acronyms that way, though the bundled list carries none.

\section{The MedWER protocol}
\label{sec:protocol}

The protocol comprises four fixed steps.

\paragraph{1. Normalization.} Both reference and hypothesis are normalized with the Whisper
\texttt{EnglishTextNormalizer} \citep{radford2023whisper,
whispernormalizer}, unmodified and held at a pinned release. Its number-word
folding step maps \emph{five hundred} $\to$ \emph{500} on both sides, so a
digit/word surface difference never scores as an error. References are
expected in written clinical form, ``500 mg'': the form a note carries and the
form the public medical corpora distribute. Strict digit/word scoring remains
available (\path{Normalizer(fold_numbers=False)}) for corpora whose numeric
surface is itself under test. Unit words are not folded: \emph{milligrams} and
\emph{mg} remain distinct tokens, so the unit convention of the references is
part of the measured surface. Normalization is Latin-script-scoped: accents fold to
ASCII via NFKD, and full-Unicode lowercasing corner cases (CJK, Greek final
sigma) are outside the guaranteed domain.

\paragraph{2. WER.} Token-level Levenshtein distance over
whitespace-delimited tokens of the normalized strings.

\paragraph{3. \medwer{}.} Both normalized sides are filtered to term-list
content, and WER is computed on what remains. The filter is a greedy
longest-match scan over the token stream, left to right. At each position it
tries candidate windows longest-first, bounded by the longest list entry. The
first match is emitted as one atomic unit, its words joined by underscore, and
its tokens are consumed; matches never overlap and the scan never backtracks.
An unmatched token is dropped and the scan advances by one.

Atomic units change the denominator, not just the edit count. Confusing
\emph{insulin glargine} with \emph{insulin lispro} is one full miss over one
reference word, not partial credit for matching \emph{insulin} while getting
the agent wrong. The same holds for longer entries: scored as three separate
tokens, \emph{3rd nerve palsy} against \emph{6th nerve palsy} would be one
substitution over three reference words; as one atomic unit it is one over
one.

\paragraph{4. Corpus aggregation.} Both metrics are micro-averaged: total
edits over total reference words. The \medwer{} denominator is the
\emph{filtered} reference length (\texttt{med\_ref\_words}), not the full one.
Samples whose \emph{normalized reference} is empty --- non-speech annotations
--- are skipped.

One case has no per-utterance value: a filtered reference that is empty while
the hypothesis carries term-list matches, such as a hallucinated drug name in
a non-medical utterance. Those matches enter the corpus numerator as
insertions and contribute nothing to the denominator, so the per-utterance
\medwer{} is undefined and only the micro-average is reported. Rates are
emitted as fractions by the tool; a rate whose denominator is zero is null,
never 0.0.

\subsection{The pinned normalizer}
\label{sec:parity}

One dependency carries behavior that can move a published number: the library
implementing the normalizer. It is pinned to an exact release, and its behavior
is asserted against a committed golden fixture file: 22 normalization cases
including accented, diacritic, and NFKD-compatibility inputs; 14 WER pairs
covering digit/word surfaces that folding reconciles and unit words it does
not; 9 \medwer{} cases exercising atomic phrases, a phrase competing with its
own first token, spelled acronyms, and
digit-bearing terms. The fixtures are regenerated only when the pin moves or
the fixture corpora change, and the suite runs on every commit, so an upstream change to normalization or to
the number parser fails the build rather than silently changing published
numbers. The term list is stored in the normalizer's own output space, every
entry a fixed point of it, so entries match normalized text verbatim.

\section{Coverage validation}
\label{sec:coverage}

A fixed denominator has a failure mode a model-derived one does not: a
missing term is silently uncounted. The list's completeness therefore needs
measurement, and measuring it against its own sources would be circular. We
validate against \textbf{BC PharmaCare's} drug data file, which lists the
drugs actually dispensed in British Columbia and was deliberately not used in
construction. The raw file is an administrative database, not a formulary: it
carries non-benefit, discontinued, and placeholder rows. We first filter to
the active benefit set as of 2026-08-19 using only the file's own fields
(296{,}144 $\to$ 34{,}112 rows; 1{,}138 generic-ingredient tokens).

Generic-name token recall is \textbf{826/1{,}138 = 72.6\% raw}. Two artifact
classes in the \emph{reference} file, not in the list, are then excluded from
the denominator: fixed-width truncations (\emph{acetaminop} for
\emph{acetaminophen}; 77 tokens) and internal abbreviations ($\leq$2 deleted
letters, same first letter; \emph{amoxiciln}; 10 tokens), giving
\textbf{826/1{,}051 = 78.6\% after cleaning}. The cleaning removes
reference-side artifacts only; the hit count is unchanged. The residual
225 tokens are dominated by discontinued agents
(\emph{chlorpropamide}, \emph{amobarbital}), excipients and herbals
(\emph{agar}, \emph{cascara}), and a thin tail of older antibiotics. Generic
recall is the coverage metric: the file's brand column mixes manufacturer and
device strings into the names, and brand-name recall (61.9\%) measures that
noise as much as the list.

\subsection{Calibration against ground-truth entity spans}
\label{sec:calibration}

String matching against a fixed list is a heuristic, and one public set lets
us calibrate it: the Eka corpus carries human-annotated medical-entity
spans, so \medwer{} can be recomputed on identical transcripts with the
annotated spans standing in for the term list. For Moonshine base the
heuristic gives 28.28\% and the span-based recomputation 24.45\%: agreement
within 4 points, with the heuristic on the conservative, higher side. Both
figures are computed over the whole Eka corpus rather than the Eka-test split
of Table~\ref{tab:baselines}; on the split the heuristic reads 30.6\%. At the
token level, the list's precision is 95.7\%. Its
recall against the full annotation breadth is 20.2\%, which is expected: the
list targets drugs, diagnoses, symptoms, and injury mechanisms in a Canadian
scope, while the annotations span the full breadth of Indian-context medical
entities, including out-of-scope Indian brand names. Recall tracks that
scope by category, from 28.8\% on clinical findings down to 5.7\% on
advice text.

\subsection{Cross-check against an NER-derived denominator}
\label{sec:nercheck}

A second check is operational: whether the fixed list changes the conclusions
a user would draw had they built the denominator with an off-the-shelf NER
model instead. We score the same four hypothesis sets on both corpora twice,
changing only the term source: the fixed list, and per-utterance entity
strings from scispaCy \citep{neumann2019scispacy}, model
\texttt{en\_ner\_bc5cdr\_md} (\textsc{chemical} and \textsc{disease} labels)
run over each normalized reference. The normalizer, atomic multi-word units,
greedy filtering, Levenshtein, and micro-averaging are held identical, so
differences are attributable to the denominator alone.

\begin{table}[ht]
\centering
\small
\begin{tabular}{lrrrr}
\toprule
& \multicolumn{2}{c}{Eka-test} & \multicolumn{2}{c}{PriMock57} \\
\cmidrule(lr){2-3}\cmidrule(lr){4-5}
Model & Fixed list & NER & Fixed list & NER \\
\midrule
Moonshine base (fp32) & 30.6 & 39.0 & 25.2 & 23.5 \\
Moonshine base (int4) & 34.4 & 42.6 & 26.9 & 26.2 \\
Whisper base.en       & 30.8 & 39.2 & 23.4 & 22.8 \\
MedASR                & 18.6 & 28.5 & 24.5 & 24.4 \\
\bottomrule
\end{tabular}
\caption{\medwer{} (\%) under the fixed term list vs.\ the NER-derived
denominator; lower is better. The fixed-list columns repeat the \medwer{}
cells of Table~\ref{tab:baselines}.}
\label{tab:nercheck}
\end{table}

On Eka-test the NER denominator is smaller, 477 reference term words against
the fixed list's 617, and every \medwer{} is higher. On PriMock57 the two
denominators are the same size (1{,}270 each) and the rates agree within 1.7
points. The question is whether either denominator changes which orderings
of two systems the corpus supports. For each of the six two-system
comparisons under each denominator, the difference in their \medwer{} was
resampled: utterances drawn with replacement, 50{,}000 replicates, one draw
shared by both systems so they are scored on identical utterances, 95\%
percentile interval on the difference. The order of two systems is supported
where that interval excludes zero; otherwise the corpus does not distinguish
them. 

On Eka-test the fixed list supports five of the six orderings and NER
four, and the two rankings agree. On PriMock57 the fixed list supports one
ordering and NER two; the rankings differ in one adjacent position, MedASR
versus Moonshine~fp32, an ordering neither denominator supports ($+0.71$
$[-2.60, +3.95]$ fixed, $-0.87$ $[-3.73, +1.99]$ NER). Moonshine~fp32 versus
Whisper is unsupported under both denominators on both corpora. No ordering
that either denominator supports is reversed by the other. The comparison
covers two test sets and four systems, so it is a
consistency check on those corpora rather than a general claim about
denominators.

\section{Baselines on open benchmarks}
\label{sec:baselines}

Table~\ref{tab:baselines} reports baselines computed with the released tool
(version 1.0.0) on two public sets. The held-out Eka test partition is 543
utterances of real accented English with real drug vocabulary; PriMock57 is
used whole and test-only: all 57 consultations, doctor channel, 3{,}405
scored utterances, conversational register. All systems are scored on greedy
recognition, no beam search and no language-model fusion; the table measures
the models under one fixed decode, and the figures are not comparable to
headline numbers from stronger decoding configurations.
References are scored as the corpora distribute them: Eka's carry written
digit forms (\emph{Dolo 650}), PriMock57's spell numbers out. Both sides are
normalized identically, so folding reconciles the two conventions rather than
charging them as errors. These are reference points for users of the tool,
not a model comparison study.

The scored set is reconstructable without redistributing the corpus. The
repository's \texttt{eval/} directory publishes the Eka-test identifiers (543
utterance ids, 410 session ids) and the frozen split recipe, plus a script that
streams the public dataset and rebuilds the reference file. The rebuilt
references reproduce the Eka columns of Table~\ref{tab:baselines} exactly.

\begin{table}[ht]
\centering
\small
\begin{tabular}{lrrrr}
\toprule
& \multicolumn{2}{c}{Eka-test} & \multicolumn{2}{c}{PriMock57} \\
\cmidrule(lr){2-3}\cmidrule(lr){4-5}
Model & WER & \medwer{} & WER & \medwer{} \\
\midrule
Moonshine base (fp32) & 22.3 $\pm$2.6 & 30.6 $\pm$4.5 & 20.2 $\pm$0.7 & 25.2 $\pm$2.8 \\
Moonshine base (int4) & 24.1 $\pm$2.7 & 34.4 $\pm$4.7 & 22.7 $\pm$0.7 & 26.9 $\pm$2.8 \\
Whisper base.en       & 23.0 $\pm$2.9 & 30.8 $\pm$4.5 & 19.1 $\pm$0.6 & 23.4 $\pm$2.7 \\
MedASR                & 17.9 $\pm$2.4 & 18.6 $\pm$3.5 & 42.8 $\pm$1.0 & 24.5 $\pm$2.8 \\
\bottomrule
\end{tabular}
\caption{Baselines through the released scorer (v1.0.0), reported here as
percentages (the tool emits fractions); lower is better. All rows greedy
decoding, no language model. Eka-test $n=543$ utterances; PriMock57
$n=3{,}405$. Intervals are 95\%, resampled over utterances with replacement
(50{,}000 replicates, percentile), summarized as $\pm$ half-width; the largest asymmetry across
cells is 0.4 points. int4 = ONNX~Runtime 4-bit block-wise weight quantization
(\texttt{MatMulNBits}; block 32, symmetric).}
\label{tab:baselines}
\end{table}

The metric surfaces three observations that overall WER hides. First, on
jargon-bearing Eka, \medwer{} runs 7.8--10.3 points above overall WER for the
general-purpose models (0.7 for MedASR). On conversational PriMock57 the gap
narrows to 4.2--5.0 points, in line with its lower medical-token density
(2.70\% against Eka-test's 11.6\%). Second, MedASR, a dictation specialist, collapses on
conversational audio by overall WER (42.8\%) yet holds the best \medwer{} on
real jargon (18.6\% on Eka, against 30.6\% for the nearest general-purpose
system): entity-restricted scoring separates domain competence from register
mismatch. Third, on Eka, quantization costs more on
the medical column than the overall column ($+3.8$ vs.\ $+1.8$ points), while
on PriMock57 the asymmetry reverses ($+1.7$ vs.\ $+2.5$): the medical
quantization penalty concentrates on jargon-dense input, and overall WER
alone shows neither effect.

\section{The tool}
\label{sec:tool}

The Python package installs with \texttt{pip install medwer}; the term list
ships inside the wheel as package data. The CLI is deterministic:

\begin{quote}
\verb|medwer score --refs refs.jsonl --hyps hyps.jsonl|
\end{quote}

It joins hypotheses to references by id; the id sets must match exactly, and
a mismatch is an error, not a silent zero. Output is one JSON object with
fixed keys: \texttt{\{utts, ref\_words, med\_ref\_words, wer,
medical\_wer\}}. The library exposes \texttt{Normalizer}, \texttt{Scorer},
\texttt{wer}, \texttt{medical\_wer}, \texttt{keep\_medical}, and
\texttt{load\_terms}. The repository additionally contains the golden
fixtures, the fixture generator, and the workflow that runs the suite on
every commit. Custom term lists are a file argument, validated at load: every
entry must be a fixed point of the scorer's normalizer. The protocol
machinery is domain-agnostic; the bundled medical list is the reference
instantiation.

\section{Limitations}
\label{sec:limits}

\begin{itemize}
\item \textbf{Scope.} The bundled list is English with a Canadian formulary
  scope; jurisdictions with different brand vocabularies
  will see lower coverage on brands, though ingredient names transfer broadly.
\item \textbf{Fixed-list breadth.} A fixed list cannot claim the entity
  breadth of an evaluation-time NER model: calibration recall against
  full-breadth annotations is 20.2\% (Section~\ref{sec:calibration}), and
  laboratory and organism vocabularies are absent for want of an open
  structured source, so terms such as \emph{hematocrit} or
  \emph{staphylococcus} appear only where an ICD-10-CM diagnosis name carries
  them. The claim is reproducibility on a declared vocabulary, not
  completeness over all medical language.
\item \textbf{Usage versus membership.} Membership is a property of the
  string, not of its use. \emph{Water} is a Drug Product Database ingredient
  (sterile water for injection) and \emph{pain} a symptom name, so each counts
  as term content wherever it appears, including in ordinary speech; 49
  single-token entries are also common English words. Filtering them out
  would delete real vocabulary; \emph{asthma}, \emph{insulin}, and
  \emph{fall} as an injury mechanism are all in that set. The list keeps
  them, and the denominator cannot distinguish the two uses. Product-line
  words that brand names contribute through the component rule
  (Section~\ref{sec:vocab}) count on the same basis.
\item \textbf{Unit surface.} Number words fold but unit words do not, so a
  reference reading ``500 mg'' and a hypothesis reading ``500 milligrams''
  count one error. The metric is therefore comparable across corpora that share
  a unit convention; references are expected in written clinical form.
\item \textbf{Normalizer provenance.} Normalization semantics, including the
  number parser, come from a pinned third-party release; the pin and the golden
  fixtures are what make a number reproducible, and moving the pin is a
  protocol change.
\item \textbf{Matching heuristic.} Greedy longest-first matching is
  deterministic but not alignment-optimal. Its deviation from an optimal
  alignment was not measured: the span calibration varies the \emph{term
  source}, not the matching rule, so the 3.8-point gap it reports bounds a
  different quantity. The filter-then-align ablation below measures the
  ordering choice on the two benchmark corpora.
\item \textbf{Filter-then-align.} Filtering both sides to term-list content
  \emph{before} alignment is a protocol-defining choice: relative to
  aligning the full strings and
  counting only edits incident on term-list tokens, it can merge a deletion
  and a distant insertion into a single substitution, or separate them.
  Holding atomicity
  constant so only the ordering varies, align-then-count reads 0.8--1.7 points
  \emph{higher} than the protocol across all eight cells of
  Table~\ref{tab:baselines}, uniformly in sign and preserving every ordering
  the protocol supports. The one ordering that flips between arms,
  Moonshine~fp32 versus Whisper~base.en on Eka-test, is one the protocol
  does not support (Section~\ref{sec:nercheck}). The protocol fixes
  filter-then-align as the defined behavior; the ablation script ships in the
  repository's \texttt{eval/} directory.
\end{itemize}

\section{Availability}
\label{sec:avail}

Code, term list, golden fixtures, and the \texttt{eval/} directory are at
\url{https://github.com/Nordis-Tech/medwer}. The code is MIT; the term list
is bare strings under Open Government Licence -- Canada attribution and US
public-domain components, with per-source provenance in the repository
\texttt{NOTICE}. The \texttt{eval/} directory carries the Eka-test
identifiers, the reconstruction script, and the filter-then-align ablation.
The Python package is on PyPI (\texttt{pip install medwer}); the v1.0.0 source
snapshot is archived at Zenodo, DOI
\href{https://doi.org/10.5281/zenodo.22103865}{10.5281/zenodo.22103865}. Every number in this paper
is scored by the released tool at version 1.0.0, which carries the
19{,}373-entry term list inside the wheel: pinning the version pins the
denominator.
What ships is the scorer, the denominator, and the fixtures. The resampling
wrapper that turns per-utterance scores from the library into intervals, and
the scripts behind the coverage validation (Section~\ref{sec:coverage}), the
span calibration (Section~\ref{sec:calibration}), and the NER cross-check
(Section~\ref{sec:nercheck}), are not part of the release. Those three
validations therefore cannot be reproduced from the published artifact alone;
their inputs are a provincial data file, licensed span annotations, and a
third-party NER model, none redistributable here.

\paragraph*{Competing interests.} MedWER is developed and maintained by
NORDIS TECH INC., which builds a commercial medical dictation product whose
accuracy is evaluated with this tool.

\bibliographystyle{plainnat}
\bibliography{refs}

\end{document}